\documentclass[conference]{IEEEtran}
\IEEEoverridecommandlockouts
\usepackage{cite}
\usepackage{amsmath,amssymb,amsfonts}
\usepackage{algorithmic}
\usepackage{graphicx}
\usepackage{textcomp}
\usepackage{xcolor}
\usepackage{booktabs}
\usepackage{url} 
\def\BibTeX{{\rm B\kern-.05em{\sc i\kern-.025em b}\kern-.08em
    T\kern-.1667em\lower.7ex\hbox{E}\kern-.125emX}}
\begin{document}

\title{GestOS: Advanced Hand Gesture Interpretation via Large Language Models to control Any Type of Robot
\thanks{Research reported in this publication was financially supported by the RSF grant No. 24-41-02039.}
}

\author{\IEEEauthorblockN{Artem Lykov}
\IEEEauthorblockA{\textit{ISR Lab} \\
\textit{Skoltech}\\
Moscow, Russia\\
artem.lykov@skoltech.ru}
\and
\IEEEauthorblockN{Oleg Kobzarev}
\IEEEauthorblockA{\textit{ISR Lab} \\
\textit{Skoltech}\\
Moscow, Russia\\
oleg.kobzarev@skoltech.ru}
\and
\IEEEauthorblockN{Dzmitry Tsetserukou}
\IEEEauthorblockA{\textit{ISR Lab} \\
\textit{Skoltech}\\
Moscow, Russia\\
d.tsetserukou@skoltech.ru}
}

\maketitle

\begin{abstract}
We present \textit{GestOS}, a gesture-based operating system for high-level control of heterogeneous robot teams. Unlike prior systems that map gestures to fixed commands or single-agent actions, GestOS interprets hand gestures semantically and dynamically distributes tasks across multiple robots based on their capabilities, current state, and supported instruction sets. The system combines lightweight visual perception with large language model (LLM) reasoning: hand poses are converted into structured textual descriptions, which the LLM uses to infer intent and generate robot-specific commands. A robot selection module ensures that each gesture-triggered task is matched to the most suitable agent in real time. This architecture enables context-aware, adaptive control without requiring explicit user specification of targets or commands. By advancing gesture interaction from recognition to intelligent orchestration, GestOS supports scalable, flexible, and user-friendly collaboration with robotic systems in dynamic environments.
\end{abstract}

\begin{IEEEkeywords}
Gesture Recognition, HRI, LLM
\end{IEEEkeywords}

\section{Introduction}

Natural and intuitive human-computer interaction (HCI~\cite{hci_review}) and human-robot interaction (HRI~\cite{hri_review}) are a critical challenges in the development of intelligent systems that operate alongside people. Among the available modalities~\cite{hci_speech_review}\cite{hci_gaze_review}, hand gestures offer an expressive and non-verbal communication channel that aligns well with human intuition and spatial reasoning~\cite{human_communication}. Yet, despite their potential, gesture-based control systems remain limited in their flexibility, generalization, and scalability—especially in dynamic or multi-agent environments~\cite{gest_recognition_review,llm_for_gesture_selection}.

\begin{figure}[h]
  \centering
  \includegraphics[width=\linewidth]{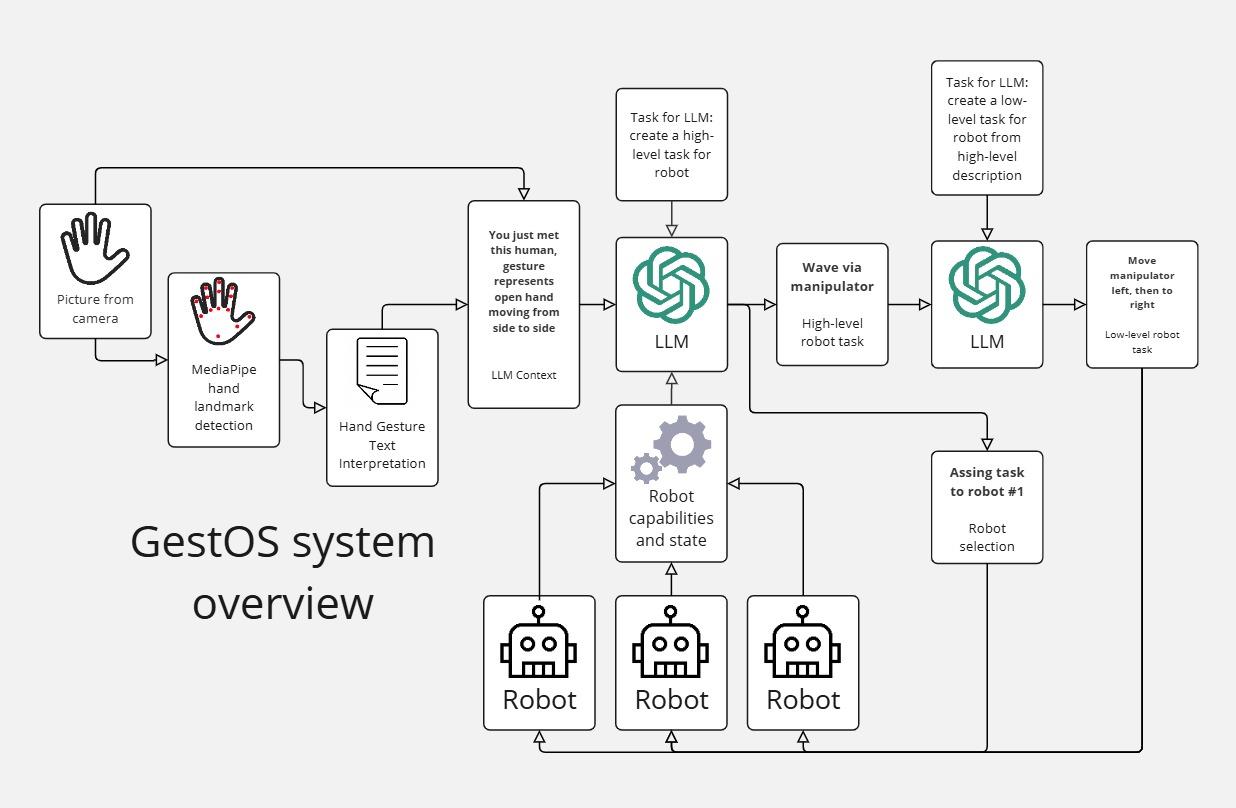}
  \caption{GestOS system overview.}
  \label{fig:arch}
\end{figure}

Traditional gesture recognition methods rely heavily on pre-defined gesture vocabularies\cite{gesture_recog_haar,rabiner,starner} and rule-based mappings~\cite{rule_based,nn_classifier_gesture}, which constrain adaptability and require retraining to support new commands or contexts. Recent advances in vision-language models (VLMs) and vision-language-action (VLA) systems have improved generalization~\cite{vlm_spatial}, but often suffer from high computational overhead, poor support for real-time execution, and limited integration with robotic control capabilities.

In this work, we introduce \textbf{GestOS}, a gesture-based operating system designed to enable high-level, flexible, and scalable control of heterogeneous robotic platforms. GestOS builds on the strengths of large language models (LLMs) for semantic reasoning while maintaining real-time responsiveness through lightweight, structured visual inputs. Specifically, GestOS uses a hand pose estimation pipeline (MediaPipe~\cite{mediapipe_paper}) to extract symbolic descriptions of hand gestures, which are then interpreted by an LLM to infer user intent.

Unlike prior systems such as GestLLM or GestureGPT that focus on gesture classification or command mapping in isolation, \textbf{GestOS advances the paradigm by introducing multi-robot task distribution as a core capability}. When a gesture is performed, GestOS reasons not only about its semantic meaning, but also about the current state, capabilities, and command interfaces of multiple available robots. It dynamically selects the most appropriate agent to execute the inferred task—whether that be a manipulator arm, mobile robot, or other device—without requiring explicit user specification. Similar ideas with heterogeneous robot control were researched is CognitiveOS paper~\cite{cognitiveos_paper}.

This architecture provides three main advantages: (1) \emph{adaptability} to previously unseen or ambiguous gestures via language-based reasoning, (2) \emph{robustness} through abstraction from raw visual input, and (3) \emph{scalability} across robot types and application domains. The overview of the system is presented in Fig.~\ref{fig:arch}.

As collaborative, multi-robot systems become increasingly prevalent in domains such as warehouse automation, assistive robotics, and smart environments, the need for intuitive and intelligent control frameworks becomes critical. By bridging gesture interpretation with multi-agent task reasoning, GestOS serves as a foundation for next-generation human-robot interfaces that are both user-friendly and context-aware.

\section{Related Work}

\subsection{Large Language Models for Robotic Control}

Recent advancements in Large Language Models (LLMs), such as GPT-4~\cite{openai2023gpt4}, PaLM~\cite{chowdhery2022palm}, LLaMA~\cite{llama}, and O1~\cite{openai_o1}, have significantly expanded the capabilities of autonomous robotic systems by enabling natural language understanding and symbolic reasoning. These models allow robots to interpret unstructured language inputs and translate them into executable high-level commands, facilitating more intuitive human-robot interaction. Notably, Huang et al.~\cite{huang2022inner} demonstrated that LLMs could be integrated into robotic control loops for step-wise reasoning in the ``Inner Monologue'' system. Similarly, SwarmGPT~\cite{swarmgpt} and FlockGPT~\cite{flockgpt} leverages LLMs for safe and flexible motion planning in drone swarms, showcasing how language-based reasoning can coordinate complex multi-agent behaviors. Furthermore, the Industry~6.0~\cite{lykov2024industry60newgeneration} paradigm envisions a new generation of industrial systems, where generative AI and swarms of heterogeneous robots collaborate to autonomously transform human ideas into finished products, closing the loop from design to deployment. However, LLM-based methods often lack grounding in visual perception and fine-grained motor control, limiting their application in real-world robotic manipulation.

\subsection{Visual Language Models for Multimodal Perception}

Visual Language Models (VLMs) such as Flamingo~\cite{alayrac2022flamingo}, LLaVA~\cite{llava_paper}, Qwen-VL~\cite{qwenvl}, DeepSeek-VL~\cite{deepseekvl} extend the capabilities of LLMs by jointly processing visual and linguistic modalities. These models have been adapted for robotics tasks like object recognition, scene understanding, and multimodal instruction following. Systems like CognitiveDog~\cite{cognitivedog_paper} and GestureGPT~\cite{gesturegpt} illustrate the potential of multimodal integration, combining gesture recognition, visual perception, and language reasoning into a unified control framework. While these approaches improve the semantic grounding of robotic systems, they face challenges in real-time deployment due to high inference latency and lack of integration with low-level control policies. In contrast, our approach addresses this gap by leveraging lightweight visual-language representations in conjunction with temporal skeleton tracking for low-latency perception-action loops.

\subsection{Vision-Language-Action Models for Embodied Intelligence}

The emergence of Vision-Language-Action (VLA) models marks a critical step toward end-to-end embodied intelligence. Unlike prior models that decouple perception from control, VLA frameworks such as RT-2~\cite{brohan2023rt2} and OpenVLA~\cite{openvla} tightly integrate high-level instruction grounding with sensorimotor execution. These models are trained on large-scale internet data and robotic experience, allowing zero-shot generalization to novel manipulation tasks. Extensions of VLA to mobile manipulation~\cite{momanipvla} and fine-grained object interaction~\cite{manipllm} underscore their flexibility across domains. Furthermore, several recent works explore the application of VLA architectures to aerial robotics, including large-scale mission generation, multi-agent code-based planning, cognitive task solving, and human-like drone racing~\cite{sautenkov2025uavvlavisionlanguageactionlargescale,sautenkov2025uavcodeagentsscalableuavmission,lykov2025cognitivedronevlamodelevaluation,serpiva2025racevlavlabasedracingdrone}. Nonetheless, VLA models remain computationally intensive and exhibit latency issues that constrain real-time interaction. Our work circumvents this limitation by coupling lightweight multimodal inference with efficient motion synthesis, thereby enabling rapid response in dynamic settings.

\subsection{Multimodal Perception in Human-Robot Interaction}

Multimodal Human-Robot Interaction (HRI) systems frequently employ pose estimation and hand landmark tracking for gesture interpretation and embodied communication. Frameworks like OpenPose~\cite{simon2017hand} and MediaPipe~\cite{mediapipe_paper} provide real-time skeleton tracking, supporting a variety of applications such as imitation learning~\cite{imitation_learning}, drone teleoperation~\cite{teleoperation,omnirace}, and versatile gesture-based control~\cite{versatileteleoperation}. Beyond gesture perception, Visual Question Answering (VQA) methods like RoboVQA~\cite{robovqa} enable scene comprehension and task guidance through natural language queries. These systems enhance semantic context understanding and object state recognition~\cite{vqabasedstaterecog}, which are essential for interactive manipulation.

\subsection{Benchmarking Multimodal Robotic Systems}

The development of benchmark datasets has been instrumental in advancing multimodal HRI. Datasets like HaGRID~\cite{hagrid} and MSR Action3D~\cite{li2010action3d} provide comprehensive video and skeleton data for evaluating gesture and action recognition systems under diverse conditions. Additionally, ALFRED~\cite{shridhar2020alfred} facilitates research on instruction following and sequential task planning in household environments. Despite their utility, current benchmarks often lack standardization across modalities and fail to represent the temporal complexity of real-world interactions. Our system builds upon these insights by introducing a unified perception-to-control pipeline designed for temporal and multimodal robustness.

\subsection{Summary}

While LLMs, VLMs, and VLAs offer promising avenues for enabling semantic reasoning, visual grounding, and embodied control in robotics, they each present specific trade-offs between flexibility, latency, and control fidelity. Moreover, integrating these models into low-latency, real-time robotic systems remains an ongoing challenge. Our work addresses these limitations by proposing a modular, multimodal control framework that combines lightweight vision-language processing with real-time pose tracking and action synthesis, enabling robust and responsive human-robot collaboration.

\section{Methodology}

\subsection{Gesture-Based Instruction Pipeline Overview}

\textit{GestOS} is a modular gesture-to-action system designed to enable real-time control of heterogeneous robotic platforms through intuitive hand gestures. Unlike prior gesture interfaces focused solely on recognition or single-agent execution, GestOS integrates perception, symbolic reasoning, and robot-aware task allocation into a unified, low-latency pipeline.

The system operates through four core stages:

\begin{enumerate}
    \item \textbf{Gesture Abstraction and Keyframe Processing}: Raw visual input is processed to extract high-confidence keyframes using MediaPipe hand tracking. These keyframes capture semantically meaningful transitions in hand pose and motion, reducing redundancy and enabling low-latency gesture summarization~\cite{keyframe}.

    \item \textbf{Semantic Feature Encoding}: Extracted hand landmarks and motion cues are transformed into structured textual descriptions, encoding information such as finger states, pointing directions, hand orientation, and spatiotemporal dynamics. This representation forms the prompt for downstream interpretation by a large language model (LLM).

    \item \textbf{Robot-Aware Task Distribution}: The inferred gesture intent is evaluated against a pool of available robots using metadata that includes hardware capabilities, current state, and supported command schemas. GestOS selects the most suitable agent to execute the task, enabling dynamic multi-robot orchestration.

    \item \textbf{Instruction Generation and Execution Memory}: A semantic reasoning module generates a task-specific command tailored to the selected robot. Validated instructions are executed, while input-output pairs are stored in a memory module to support few-shot generalization and user-specific adaptation in future interactions.
\end{enumerate}

This architecture is designed for robust performance under real-world conditions, supporting generalizable gesture control without explicit per-gesture training or hardcoded command mappings. The modularity of GestOS allows seamless extension to new robot types, gesture sets, or interaction paradigms, making it well-suited for scalable, multimodal human-robot interaction.

\subsection{Gesture Abstraction and Keyframe Processing}

To minimize computational cost while preserving gesture fidelity, the pipeline begins with real-time frame sampling and keyframe selection. Video input is first processed using the MediaPipe framework for hand landmark estimation, producing a low-dimensional but semantically rich representation of hand pose. 

Only frames with high model confidence are retained. From this filtered stream, keyframes are extracted using a hybrid strategy that combines:
\begin{itemize}
    \item \textbf{Motion-based detection}, identifying frames with significant pose variation and hand center frame-to-frame distance.
    \item \textbf{Hand-state transitions}, tracking changes in hand openness, orientation, and finger groupings.
\end{itemize}

This approach reduces temporal redundancy while preserving meaningful spatiotemporal dynamics. These keyframes serve as anchors for subsequent semantic encoding.

\subsection{Semantic Feature Encoding}

Keyframes alone are insufficient for large language models (LLMs), which are not inherently capable of interpreting raw image or coordinate data without domain-specific training. Therefore, extracted landmark information is converted into structured textual descriptions.

Each gesture is abstracted using the following features:
\begin{itemize}
    \item Binary state of each finger (e.g., in/out of fist)
    \item Directional encoding (e.g., “index finger pointing left-up”)
    \item Finger groupings based on spatial proximity
    \item Hand orientation relative to the camera
    \item Trajectory information between keyframes
\end{itemize}

This symbolic representation serves as the prompt for the LLM, which then interprets the gesture semantically, identifying its potential communicative or operational meaning.

\subsection{Gathering Robotic Requirements for Task Distribution}

A key innovation of \textit{GestOS} lies in its ability to dynamically allocate tasks across heterogeneous robots based on real-time feasibility and contextual suitability. To accomplish this, the system maintains and queries a structured repository of robot-specific metadata that guides the task distribution process.

Each robot in the system is described using three complementary metadata categories:

\begin{enumerate}
    \item \textbf{Robot Descriptions}: Static profiles detailing each robot's hardware characteristics, degrees of freedom, manipulation or mobility capabilities, and supported sensors.
    
    \item \textbf{Command Schemas}: Enumerations of executable commands or skills unique to each robot, formatted in a machine-readable schema for compatibility checking.
    
    \item \textbf{Live State Feedback}: Real-time data capturing each robot’s operational status, including sensor health, joint availability, current task load, and spatial location (if applicable).
\end{enumerate}

When a gesture is semantically interpreted as a high-level task, GestOS cross-references this intent with the capabilities and current status of all registered agents. It then selects the most appropriate robot to fulfill the task, prioritizing feasibility, efficiency, and task-context alignment.

This decision-making process forms the bridge between abstract gesture interpretation and grounded execution. It enables GestOS to move beyond fixed gesture-command bindings and toward context-sensitive orchestration of multiple robots—supporting adaptive, scalable, and safe interaction in multi-agent settings.

\subsection{Instruction Generation and Memory-Augmented Execution}

In the final stage of the pipeline, the system combines the semantically interpreted gesture with the selected robot's capabilities to generate an executable instruction. This is performed by a large language model (LLM), which uses structured gesture descriptors and robot metadata as contextual input to produce a task-specific command.

To enhance reliability and alignment with robot-executable primitives, this process is decomposed into the following subtasks:

\begin{itemize}
    \item \textbf{Semantic Parsing}: Identification of the gesture’s high-level meaning, including intent classification and labeling (e.g., “pick up,” “turn left”).
    
    \item \textbf{Task Formulation}: Conversion of the parsed intent into a task description that aligns with the robot’s domain.
    
    \item \textbf{Instruction Matching}: Mapping the formulated task onto supported command schemas using a classifier or rule-based filter.
\end{itemize}

If a direct match is found, a structured command is issued to the robot. If the task is unsupported or ambiguous, an \emph{explainer module} attempts to decompose it into valid subtasks or fallback commands. This promotes graceful degradation and increases coverage in open-ended interactions.

To support personalization and long-term learning, all successful input–output pairs (gesture description, selected robot, command) are embedded and stored in a vector-based memory module. This memory enables few-shot prompting for future interactions, improving responsiveness to recurring tasks, user-specific styles, and evolving robot capabilities.

This memory-augmented architecture balances semantic generalization with execution fidelity, enabling robust and flexible gesture-based control in dynamic, multi-robot environments.


\section{Experimental Results}

\subsection{Objective}

The goal of this evaluation is to assess the command recognition accuracy of \textit{GestOS} across two heterogeneous robotic platforms: a robotic manipulator (UR3) and a quadruped robot (Unitree GO1, referred to as "robodog"). The experiment tests whether GestOS can accurately interpret free-form hand gestures and map them to appropriate high-level commands without relying on predefined gesture-command bindings.

\begin{figure}[h]
  \centering
  \includegraphics[width=\linewidth]{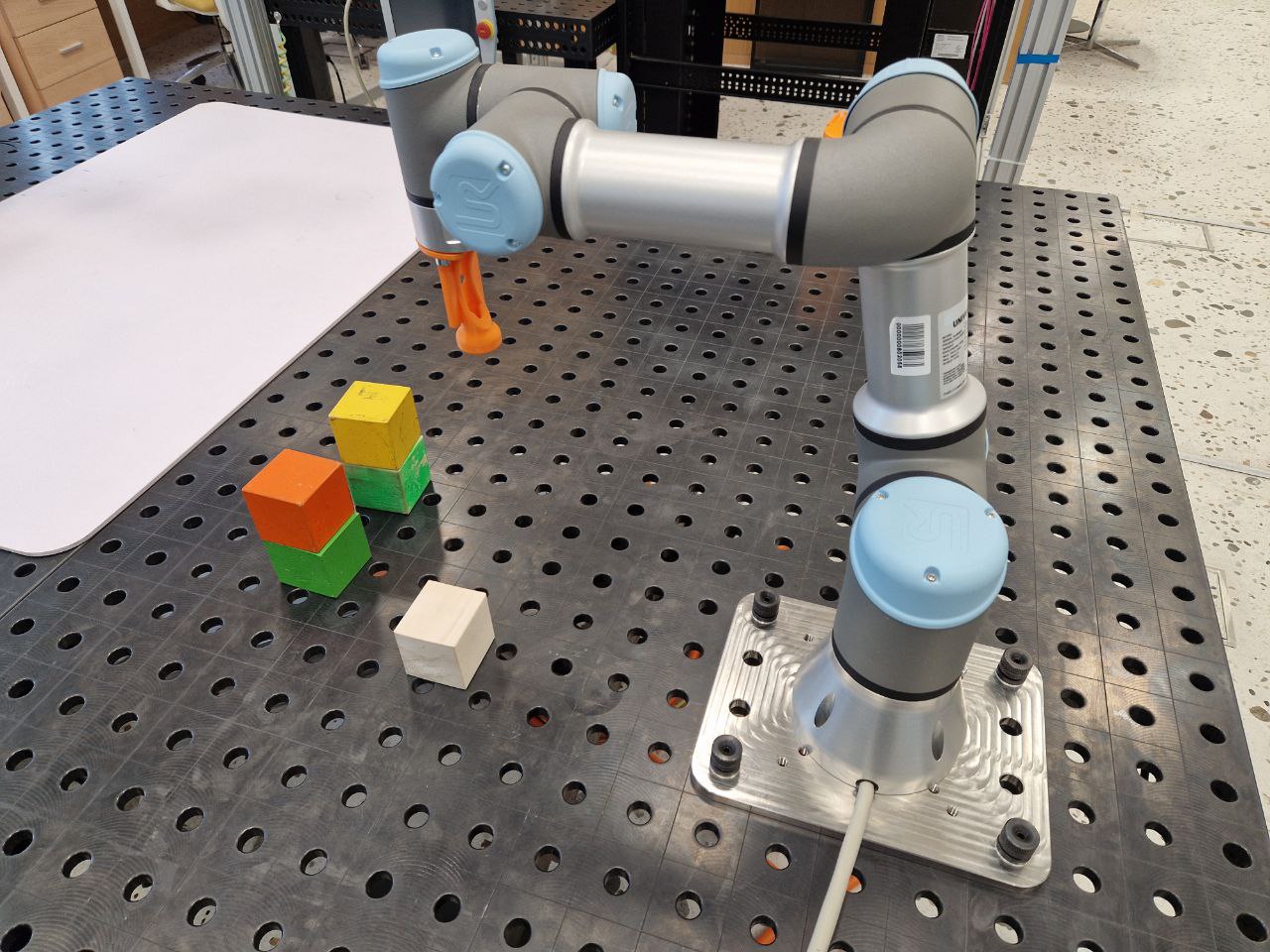}
  \caption{Universal Robots UR3 manipulator}
  \label{fig:ur3}
\end{figure}

\subsection{Robotic Platforms and Command Sets}

Two robots were used, each with its own distinct command set:

\textbf{Manipulator (UR3):}
\begin{itemize}
    \item \texttt{manipulator\_high\_five} – Give a high five to the user
    \item \texttt{manipulator\_select\_left\_item} – Select item positioned to the left
    \item \texttt{manipulator\_select\_right\_item} – Select item positioned to the right
    \item \texttt{manipulator\_turn\_object\_around} – Rotate object to show all sides
    \item \texttt{manipulator\_close\_gripper} – Grasp object
    \item \texttt{manipulator\_open\_gripper} – Release object
\end{itemize}

\vspace{0.5em}
\textbf{Robodog (GO1):}
\begin{itemize}
    \item \texttt{robodog\_give\_paw} – Lift front leg to simulate a paw shake
    \item \texttt{robodog\_stand\_up} – Rise on command when user gestures upward
    \item \texttt{robodog\_stand\_down} – Sit or lie down
    \item \texttt{robodog\_come\_closer} – Approach the user on beckoning gesture
    \item \texttt{robodog\_wagging\_tail} – Simulate tail-wagging behavior
\end{itemize}

\subsection{Procedure}

Each robot was tested with a set of gesture inputs that semantically correspond to the listed commands. For each gesture, the following steps were performed:

\begin{enumerate}
    \item The operator performed a gesture in view of the perception system.
    \item GestOS processed the gesture through the full pipeline (keyframe abstraction, semantic encoding, robot selection).
    \item A command was generated and matched to the predefined set of supported commands for the target robot.
    \item The command was validated and executed if feasible.
\end{enumerate}

\subsection{Metric: Command Recognition Accuracy}

Performance was quantified using the \textbf{command recognition accuracy}, defined as the percentage of trials where the correct command was inferred from the gesture. Table~\ref{tab:command_accuracy} reports results for all commands across both robots.

\begin{table}[h]
\centering
\caption{Command Recognition Accuracy (\%) for Each Robot and Command}
\label{tab:command_accuracy}
\begin{tabular}{lcc}
\toprule
\textbf{Command} & \textbf{Robot} & \textbf{Accuracy (\%)} \\
\midrule
manipulator\_high\_five          & UR3       & 89 \\
manipulator\_select\_left\_item  & UR3       & 98 \\
manipulator\_select\_right\_item & UR3       & 96 \\
manipulator\_turn\_object\_around & UR3       & 83 \\
manipulator\_close\_gripper      & UR3       & 72 \\
manipulator\_open\_gripper       & UR3       & 81 \\
\midrule
robodog\_give\_paw               & GO1       & 95 \\
robodog\_stand\_up               & GO1       & 68 \\
robodog\_stand\_down             & GO1       & 96 \\
robodog\_come\_closer            & GO1       & 85 \\
robodog\_wagging\_tail           & GO1       & 61 \\
\bottomrule
\end{tabular}
\end{table}

\subsection{Observations}

GestOS demonstrated strong command recognition performance across both platforms, with the majority of commands achieving over 80\% accuracy. The UR3 manipulator showed particularly high recognition rates for spatial selection gestures such as \texttt{select\_left\_item} (98\%) and \texttt{select\_right\_item} (96\%), indicating reliable interpretation of directional intent.

However, several action-related commands exhibited lower accuracy. For example, \texttt{manipulator\_close\_gripper} (72\%) and \texttt{open\_gripper} (81\%) underperformed relative to others. Similarly, \texttt{robodog\_come\_closer} achieved 85\% accuracy—slightly lower than expected. All three involve dynamic, multi-phase gestures with temporal motion, which pose challenges for both the human operator (in gesture consistency) and the LLM (in temporal intent parsing). These results suggest that compound or time-dependent gestures are a current limitation of the pipeline.

Another source of confusion stems from semantic overlap in gesture meanings. Commands such as \texttt{robodog\_wagging\_tail} (61\%), \texttt{robodog\_stand\_up} (68\%), and \texttt{manipulator\_high\_five} (89\%) share upward hand motion or greeting-related cues. This conceptual and spatial proximity likely increases misclassification risk, particularly for LLMs inferring intent from similar symbolic representations.

Overall, GestOS effectively interprets free-form gestures for both spatial selection and social interaction tasks, but performance degrades when handling dynamic gestures or semantically adjacent commands. These insights motivate future improvements such as dynamic gesture segmentation, temporal embeddings, and real-time user feedback to resolve ambiguity during execution.


\section*{Conclusion}

This work presented \textbf{GestOS}, a gesture-based operating system that advances LLM-driven interaction frameworks by enabling multi-robot task distribution and context-aware command generation. In contrast to prior systems that rely on static gesture-command mappings or focus solely on single-agent control, GestOS dynamically reasons over real-time robot states, hardware capabilities, and supported instruction sets to assign tasks intelligently across heterogeneous platforms.

Experimental validation on a robotic manipulator and a quadruped robot demonstrated that GestOS can semantically interpret free-form gestures and issue accurate, robot-specific commands without relying on hardcoded mappings. The system achieved particularly high accuracy on spatial and symbolic gestures, highlighting its strength in generalizable, real-time multimodal control. Performance was more variable for dynamic or semantically overlapping gestures, revealing challenges in both user consistency and LLM-based intent resolution—an area that motivates further improvement.

Future work will focus on supporting more diverse robot types, enabling richer and compound gestures, and incorporating temporal modeling and real-time user feedback to improve command disambiguation. These developments aim to position GestOS as a robust, scalable foundation for intuitive human-robot collaboration in increasingly complex environments.



\end{document}